# A Mechanistic Explanatory Strategy for XAI

**Marcin Rabiza**

**Abstract:** Despite significant advancements in XAI, scholars note a persistent lack of solid conceptual foundations and integration with broader scientific discourse on explanation. In response, emerging research draws on explanatory strategies from various sciences and the philosophy of science literature to fill these gaps. This paper outlines a mechanistic strategy for explaining the functional organization of deep learning systems, situating recent developments in explainable AI within a broader philosophical context. According to the mechanistic approach, the explanation of opaque AI systems involves identifying mechanisms that drive decision making. For deep neural networks, this means discerning functionally relevant components, such as neurons, layers, circuits, or activation patterns, and understanding their roles through decomposition, localization, and recomposition. Proof-of-principle case studies from image recognition and language modeling align these theoretical approaches with mechanistic interpretability research from OpenAI and Anthropic. The findings suggest that pursuing mechanistic explanations can uncover elements that traditional explainability techniques may overlook, ultimately contributing to more thoroughly explainable AI.

## 1 Introduction

In recent years, *deep learning* has emerged as the dominant approach in *artificial intelligence* (AI). Due to their inherent complexity, deep learning models are often referred to as "black boxes," making it challenging to determine how specific inputs lead to predictions. To address this, researchers have turned to the *explainable AI* (XAI) research program, which seeks to enhance transparency, interpretability, and explainability by validating the decision-making processes of opaque AI systems and making model behavior understandable to stakeholders with various epistemic needs.[1]

M. Rabiza
Institute of Philosophy and Sociology, Polish Academy of Sciences, Warsaw, Poland
Institute for Philosophy, Leiden University, Leiden, The Netherlands
e-mail: marcin.rabiza@gssr.edu.pl
ORCID: 0000-0001-6217-6149

Despite notable technical advancements in XAI (e.g., Linardatos et al., 2021; Montavon et al., 2018), scholars note a persistent lack of solid conceptual foundations and integration with broader discourse on scientific explanation. Many current XAI methodologies excel in generating localized explanations but fall short of providing a holistic understanding of processes governing AI systems operation (Kästner & Crook, 2024; Mittelstadt et al., 2019). This limitation appears critical for mitigating both technical and ethical risks associated with high-stakes decision making. Moreover, AI explanations have predominantly followed a technology-focused approach, disregarding established theories and empirical insights from philosophy and non-technical disciplines, leaving a substantial research space underexplored (Miller, 2019).

In response, a new wave of fundamental XAI research is now drawing on classical philosophical approaches to explanation (e.g., Beisbart & Räz, 2022; Erasmus et al., 2021; Lipton, 2018; Miller et al., 2017; Mittelstadt et al., 2019; Páez, 2019; Watson & Floridi, 2020; Zednik, 2021; Zerilli et al., 2019). Within this emerging body of work, there is a growing consensus that XAI should move beyond correlation-based methods and input–output analyses toward examining AI systems from a functional perspective, with greater emphasis on causal reasoning and interventionist methodologies (Kästner & Crook, 2024; Marrow et al., 2020; Millière & Buckner, 2025).

This paper outlines a *mechanistic strategy* for explaining the functional organization of deep learning systems, situating recent advancements in XAI methods— especially in mechanistic interpretability—within a broader philosophical context. To this end, I draw upon the tradition of the *new mechanism* in the philosophy of science while leveraging examples derived from XAI engineering practices. The mechanistic approach, drawing from causal and interventionist schools of thought, has historically been both dominant and highly successful in fields such as biology, neuroscience, and cognitive science. Given the biological inspiration behind deep learning research, there is no principled reason why such an approach cannot be similarly fruitful in XAI.

The structure of the paper is as follows. Section 2 introduces the concept of a mechanism based on a minimal neomechanistic theory of explanation and argues for its applicability to XAI. Section 3 presents a mechanistic interpretation of deep learning, explaining the operations of deep neural networks by identifying decision-making mechanisms through discovery heuristics of decomposition, localization, and recomposition. Proof-of-principle case studies from image recognition and language modeling are used to align these theoretical approaches with recent mechanistic interpretability research from OpenAI and Anthropic. Section 4 considers a potential objection to the philosophical characterization of deep neural networks as mechanisms. Section 5 discusses the epistemic significance of the mechanistic approach for XAI. Section 6 concludes.

---

[1] *Interpretability* is typically defined as "the degree to which an observer can understand the cause of a decision" (Miller, 2019, p. 8). Explanation, therefore, is one mode through which an observer may obtain such understanding. In the machine learning literature, the terms "interpretability" and "explainability" are often used interchangeably (Molnar, 2022, §3)—a convention I will follow for now until I introduce a specific understanding of interpretable and comprehensible systems from Doran et al. (2017) later in the paper.

## 2 Neomechanistic Theory of Explanation

In philosophical discourse on scientific explanation, one prominent approach is characterized by the neomechanistic theory of explanation, which emphasizes the logic of "explaining *why* by explaining *how*" (Bechtel & Abrahamsen, 2005, p. 422). According to the new mechanists, explaining why something happens involves identifying the underlying mechanisms that give rise to observed behavior. *Mechanisms* are identified by the phenomena they produce (Illari & Williamson, 2012), their functional roles (Machamer et al., 2000), and their operating "parts" and "interactions." (Bechtel & Abrahamsen, 2005). For example, Glennan (1996, p. 52) defines a mechanism as "a complex system which produces that behavior by the interaction of a number of parts according to direct causal laws."

According to Machamer et al. (2000, p. 3), mechanisms are identified and individuated by the "activities" and "entities" that constitute them, as well as their functional roles and setup and termination conditions. *Entities* are components or parts defined by their properties, such as location, structure, and orientation, that engage in activities based on specific characteristics. *Activities* are temporal producers of change, characterized by aspects such as spatiotemporal location, rate, duration, types of entities involved, and other operational properties. The roles that entities play through their activities are considered their functions within the mechanism. The specific organization of these elements determines how collectively they produce the observed phenomenon (Machamer et al., 2000).

Mechanistic explanation proceeds by characterizing the phenomenon under study and identifying the mechanisms responsible for it. According to Bechtel and Abrahamsen (2005), to explain a mechanism, scientists must pinpoint its components, understand the functions these parts perform, and determine their organization to produce the phenomenon. This process relies on scientific discovery methods, incorporating heuristic strategies such as decomposition and localization (Bechtel & Richardson, 2010).

*Decomposition* involves breaking the mechanism into its structural or functional components. Structural decomposition focuses on the physical aspects of parts, such as size or shape, while functional decomposition looks at the parts' roles, causal powers, and overall contributions to the mechanism (Piccinini & Craver, 2011). Functional decomposition assumes that the system's behavior results from its subfunctions. Structural decomposition further breaks down these functions into their physical components. This process starts with hypothetical component parts, refining the breakdown as more is learned about the system's operations, with both types of decomposition eventually integrating to form a complete explanation.

*Localization* complements the decomposition process by mapping component operations onto their respective parts. While decomposition involves breaking down the mechanism into parts and operations, localization identifies activities from the task decomposition and links them to component behaviors or capabilities (Wright & Bechtel, 2007). Sometimes, physical components can be directly identified, but often their existence is inferred using functional tools without direct observation. Bechtel and Abrahamsen (2005) note that even inferred components are treated as essential parts of the mechanism. Localization involves a genuine commitment to the functions identified in the task decomposition and the use of appropriate methods to demonstrate that something within the system is performing each of these functions.

Mechanistic explanations describe the relevant entities, their properties, and the activities that connect them, by demonstrating how actions at one stage influence those at successive stages. Glennan (2017) notes that mechanistic models can be depicted through diagrams accompanied by linguistic explanations. These diagrams typically illustrate spatial relations and structural features of the mechanism, with related activities depicted as labeled arrows (see Fig. 2.1). Although the basic arrangement of a mechanism might be linear, it can also include more complex structures like forks, joins, cycles, and non-linear interactions.

**Fig. 2.1** A diagrammatic representation of a mechanism (reproduced from Craver, 2007). At the top is the phenomenon: some system *S* engaged in behavior $\psi$. Beneath it are the parts (the *X*s) and their activities (the $\varphi$s) organized together.

As Figure 2.1 might suggest, mechanisms form nested, multilevel hierarchies, in which lower-level entities and activities serve as components for higher-level phenomena, effectively becoming mechanisms themselves. These hierarchies have a finite structure and do not decompose indefinitely; the lowest level of description is determined by practical considerations and stakeholder interests. Although all mechanisms are ultimately based on fundamental, non-causal physical laws, seeking explanations at this level is typically neither practical nor beneficial. Instead, explanations tend to bottom out at components that are fundamental or unproblematic within a specific scientific discipline or explanatory practice (Machamer et al., 2000).

In the optional stage of mechanism discovery, as described by Bechtel and Abrahamsen (2013), scientists may *recompose* what they have learned about the functional parts into an explanatory model, such as a mathematical or computational model. The goal is to create a model from a catalog of entities likely to be causally relevant to a phenomenon based on their identified internal division of labor.

The *mechanistic explanatory strategy* that emerges from this description resembles something like a *causal narrative*, in the sense that it outlines sequences of events involving entities interacting with each other, illustrating how their spatial and temporal arrangement produces or sustains the explanandum (Glennan, 2017). An important question at hand is: How might this framework be adapted to analyze and explain opaque AI systems?

# 3 Mechanistic Interpretation of Deep Learning

## 3.1 Mechanistic Structure of Deep Neural Networks

The mechanistic explanatory strategy has been adopted across various scientific fields, being especially prevalent in neuroscience (Kostić & Halffman, 2023). Given the biological inspiration of *deep neural networks* (DNNs), and certain computational parallels to the brain's processes (Schyns et al., 2022), applying mechanistic principles to opaque AI systems also holds considerable promise. Theorists like Kästner and Crook (2024) argue that as modern AI systems grow increasingly complex, they should be analyzed through the same prism as biological organisms—that is, with a focus on their functional organization. In this vein, a growing number of XAI techniques aim at *model explainability* by investigating the internal mappings of DNNs through interventionist methods (see an overview in Millière & Buckner, 2025).

Building on these points, the following sections argue that applying a mechanistic approach to DNNs can significantly enhance our understanding of their internal structures and functional organization, thereby addressing some of the key desiderata of XAI. By systematically deciphering the computational and representational architecture of deep learning models, this approach enables *model-driven explanations* that illuminate how specific computations give rise to individual decisions. As a result, it supports improved prediction and control over target AI systems. Mechanistic principles may provide a foundation for a coordinated research strategy that extends beyond isolated techniques, fostering a more holistic and systematic understanding of complex AI models (cf. Kästner & Crook, 2024).

In the neomechanistic framework, explaining AI systems entails identifying the mechanisms behind their decision-making processes using discovery heuristics of decomposition, localization, and recomposition. The goal is to understand the properties driving behavior and how these are orchestrated through component interactions (see Fig. 3.1). By dissecting neural networks into comprehensible components, researchers can better grasp each part's function and structure, gaining insights into the network's overall behavior.

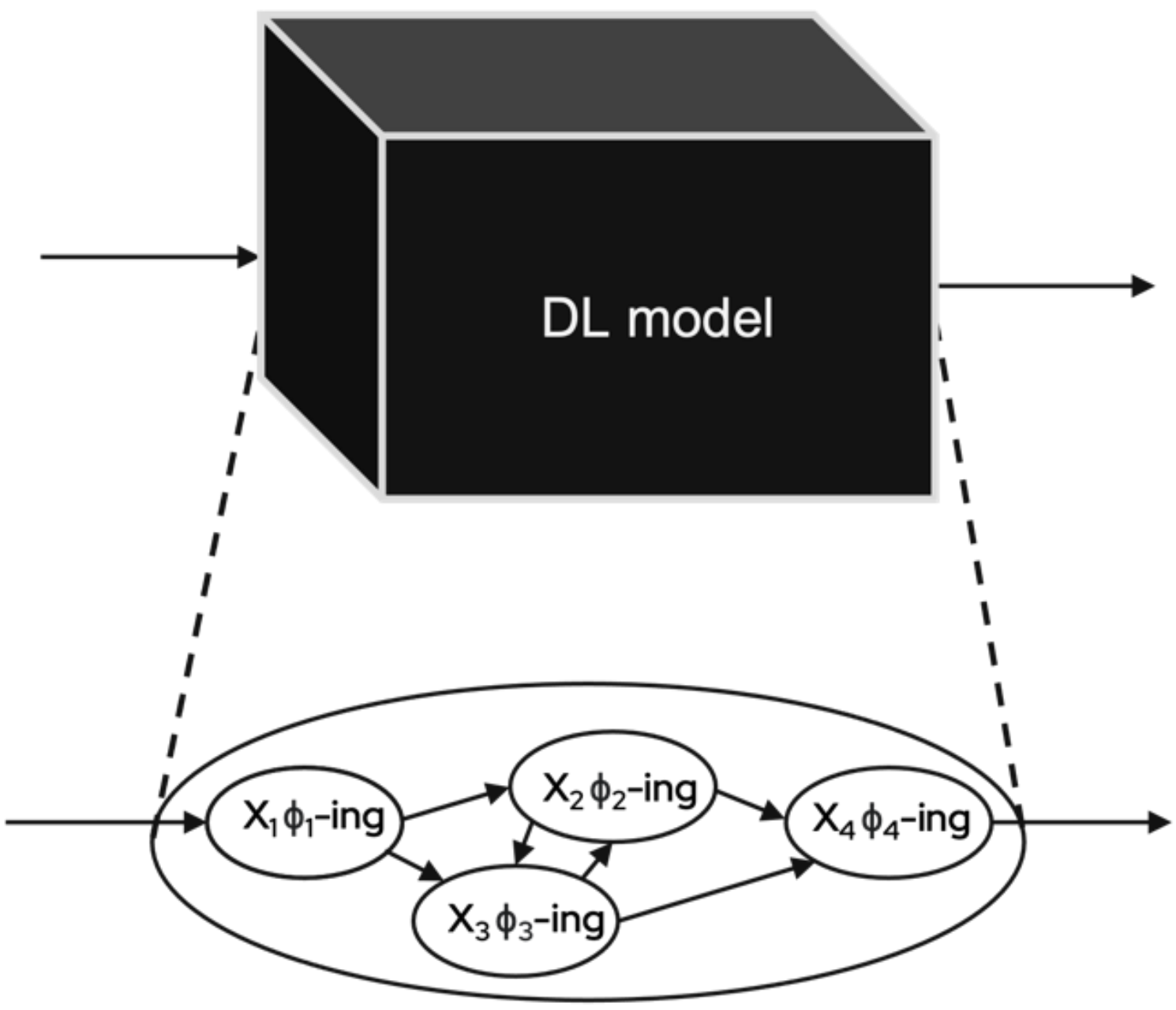

**Fig. 3.1** Schematic representation: internal structure of the mechanism of the deep learning model analyzed in the neomechanistic framework.

The first step involves identifying the correct components for analysis—the candidates for the AI system's *epistemically relevant elements* (EREs; Humphreys, 2009). Neurons, while fundamental computational units of neural networks, often fall short as effective units for human interpretation. Despite performing straightforward arithmetic, individual neuron-like units viewed in isolation fail to clearly demonstrate how their functions contribute to the network's overall behavior. Consequently, researchers look to other network components as potential EREs: robust, interpretable patterns that sustain system behavior and support explanatory aims (Kästner & Crook, 2024).

This search is compatible with deep learning because DNNs are composed of causally and functionally relevant components. These include entities (medium-independent vehicles) and activities (manipulations these vehicles undergo), which are central to the computational mechanisms of deep learning. Regardless of architectural variation, deep networks share common elements such as neurons, connections, filters, circuits, or features. Although DNNs are implemented as multidimensional arrays of numbers with vectorized operations in programming languages like Python, computer scientists point out that "neural network parameters are in some sense a binary computer program which runs on one of the exotic virtual machines we call a neural network architecture" (Olah, 2022). Within this environment, mechanistically relevant components are, in principle, identifiable.

During training, these components participate in activities such as activation, backpropagation, and error minimization, often triggered when incoming signals exceed learned thresholds. Through these interactions, deep networks develop specialized functional roles—frequently unanticipated by designers—that collectively generate the system's behavior. This emergent organization underpins the possibility of mechanistic part–whole decomposition.

A clear example of such "mechanistic compatibility" is found in deep *convolutional neural networks* (CNNs), widely used in image recognition and computer vision. During training, a CNN processes two-dimensional labeled images to establish weights that encode extracted data patterns. It employs organized entities such as neurons, larger neuronal circuits, convolutional kernels (filters), or various kinds of layers, along with activities like convolutions, ReLU and softmax activations, and pooling. Together, these components orchestrate the mechanisms of feature extraction and image classification (see Fig. 3.2).

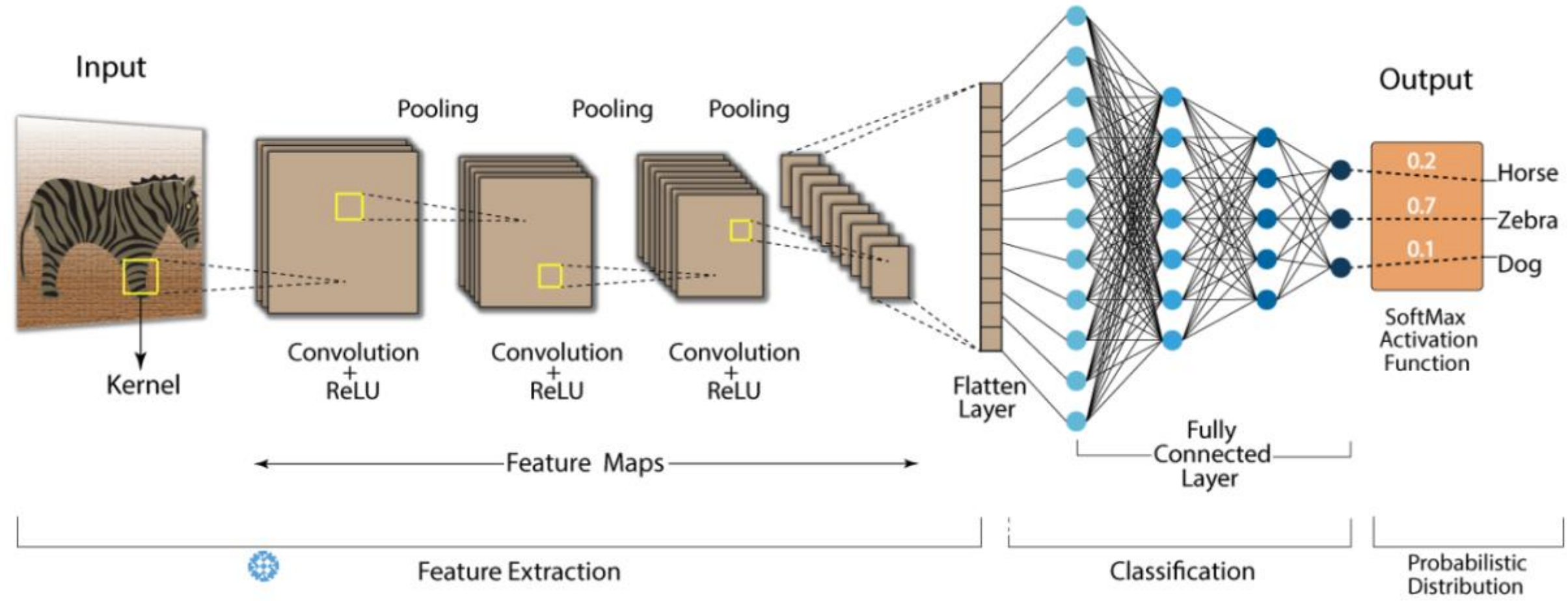

**Fig. 3.2** Architecture of a deep convolutional neural network (reproduced from Shahriar, 2023).

The inference phase of image recognition begins with *setup conditions* that include initializing model parameters. An input image is then introduced and subjected to a series of transformations across multiple layers to extract and refine features. *Intermediate activities* of this process involve convolution, activation, and pooling. Convolutional kernels slide across the input signal, detecting features similar to biological neural networks' receptive fields and generating feature maps for subsequent layers. ReLU activations eliminate negative values, activating only nodes that exceed a certain threshold. Pooling layers reduce data dimensionality by merging outputs from neuron clusters into single neurons, using hierarchical patterns to evolve simpler features into more complex ones.

As processing continues, layers represent diverse image features such as edges, lines, and curves, with higher layers capturing more complex, "meaningful," and abstract shapes. The process culminates in fully connected layers, where the softmax function transforms raw outputs into class probability scores, marking the *termination condition* of the inference phase.

Under stable and consistent conditions, this sequence exhibits the deterministic regularity characteristic of genuine mechanisms.

In this setup, CNN mechanisms form multilevel hierarchies, in which lower-level entities and activities enable higher-level phenomena, illustrating the mechanistic structure of their internal organization. For example, input convolution is crucial for feature mapping, which, through iterative application, integrates into the network- level mechanism that yields image classification in the fully connected layers.

## 3.2 Implementation of Mechanism Discovery Heuristics

Having explored the mechanistic structure of DNNs, the next step is to consider how decomposition and localization heuristics can be systematically applied to identify and understand EREs within these networks. In XAI engineering practice, these strategies can be implemented through established analytical techniques tailored to specific applications.

In computer vision, for example, input heatmapping and feature visualization techniques can be used to generate saliency maps that highlight specific pixels or regions most predictive of the model's output. Such visualizations clarify the operations performed by components across layers, aiding functional decomposition. Common methods include activation maximization, regularized optimization, network inversion, deconvolutional neural networks, network dissection-based visualization, or layer-wise relevance propagation (Montavon et al., 2018; Qin et al., 2018; Yosinski et al., 2015). Such techniques reveal how each network layer transforms inputs into progressively more abstract and meaningful representations, enabling researchers to decompose complex mechanisms like face recognition into simpler feature detectors for elements like ears, eyes, or noses. An example of hierarchical feature representations processed within a CNN is shown in Fig 3.3.

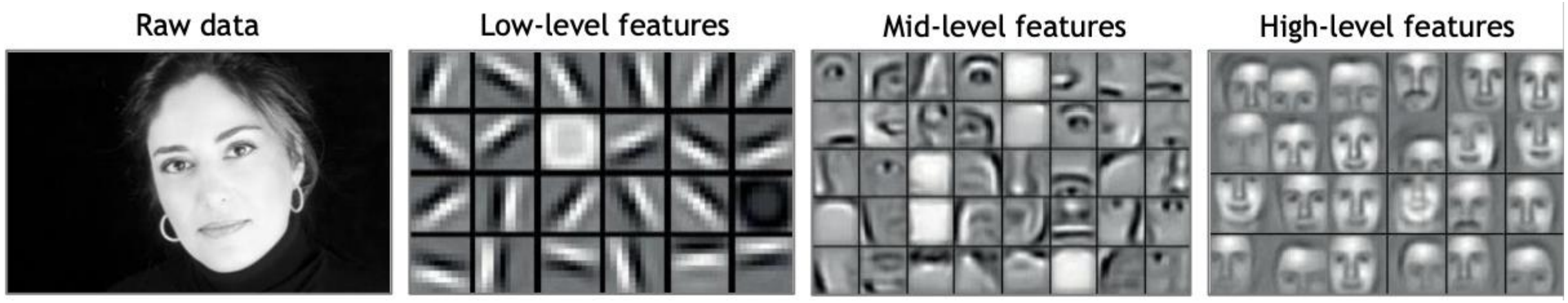

**Fig. 3.3** Visualization of features across layers of a CNN for input images of faces (reproduced from Karmakharm, 2018).

While saliency methods do not, on their own, provide complete mechanistic explanations, they play an important role in decomposing DNNs in image recognition tasks. By detecting distributed structures and activation patterns that extend beyond individual neurons, these methods help identify potential EREs, provided the detected patterns genuinely represent features. Such criteria can often be satisfied through domain-specific intervention techniques, as recent work in mechanistic interpretability has demonstrated.

*Mechanistic interpretability* aims to reverse-engineer neural networks by examining their internal structure and identifying computational mechanisms—from learned weights to human-interpretable algorithms—thereby grounding causal and functional claims about deep learning models. Unlike approaches that rely solely on correlation-based metrics at the output level, mechanistic interpretability traces the full computational process from inputs to outputs. Central to this approach is the identification of two key types of representational and structural elements: features (human-interpretable properties and semantic relationships in data) and circuits (subparts of a model that convert one set of features into newly interpretable forms). By analyzing which components activate in response to particular features, researchers can reconstruct the functional architecture of the network and decompose it into hierarchically organized structures.

Methodologically, mechanistic interpretability employs a spectrum of analytical, observational, and interventionist techniques to uncover causal roles within deep learning architectures.[2] This structured approach connects abstract algorithmic functions with concrete mechanistic implementations, yielding a distinctive form of explanation for complex AI systems. Prominent organizations such as OpenAI, Anthropic, Google, Redwood Research, ARC, and Conjecture have advanced this work, recognizing its importance for XAI, especially in AI safety contexts (Bricken et al., 2023a; Cammarata et al., 2021; Chan et al., 2022; Christiano, 2022; Conmy et al., 2023; Cunningham et al., 2023; Elhage et al., 2021; Olah et al., 2017; Olsson et al., 2022; Schwettmann et al., 2023).

A representative example is the study by Cammarata et al. (2021) at OpenAI, a which seeks to reverse-engineer image classification circuits into explicit, human-understandable explanations and then reimplement the inferred algorithms in a new DNN from scratch. Their analysis concentrates specifically on a curve-detecting circuit in the fifth convolutional layer of the InceptionV1 model. Such circuits are not predefined architectural modules; they emerge during training as functional units formed through the self-organization of neurons within the network's learned structure.

[2] Notably, several causal intervention methods for interpreting large language models have recently been examined in the philosophical literature, including work by Kästner and Crook (2024) and by Millière and Buckner (2025).

To understand the functional organization of these circuits (the "mechanics of curve detectors," as the authors describe it), Cammarata et al. follow decomposition and localization strategies. Using *decomposed feature visualization*, they construct grids that display amplified weights from upstream layers to a downstream neuron of interest. By iteratively applying this technique across neurons, labeling and grouping them, the authors categorize neurons in the first five convolutional layers of InceptionV1 into layer-specific "families" that together constitute the curve-detection mechanisms.

After identifying curve detectors, the researchers trace their connections to determine how upstream neurons influence their activity, visualizing the relevant connection weights. Extending this analysis back to the input layer yields a detailed picture of circuit-level interactions and supports classifying the circuit as a mechanistically meaningful ERE distinct from the surrounding network. The authors then construct a high-level "circuit schematic" that maps the principal components and their organization, yielding a clear explanatory narrative: "Gabor filters evolve into proto-lines, which assemble into lines and early curves, ultimately forming curves" (Cammarata et al., 2021).

Leveraging insights from this discovery process, the researchers then recompose the curve detection mechanism of InceptionV1 by manually configuring the weights of a blank DNN to replicate the identified neuron families and circuit interactions. They compare the behavior of this manually constructed network with InceptionV1 using identical synthetic stimuli, and analyze the responses with feature visualization and related XAI techniques. The experiments showed that the artificially recomposed curve detectors closely resemble those that emerged during training. This result indicates that the functional decomposition of part of InceptionV1 was successful and accurately reflects the mechanistic organization of the curve detection circuit.

Anthropic, known for developing Claude (a large language model comparable to ChatGPT from OpenAI), has adopted similar decomposition and localization strategies to break down language models into interpretable, structurally distinct functional components. Noting that individual neurons are often unsuitable as units of analysis due to their *polysemanticity*, Bricken et al. (2023a, 2023b), use a *sparse autoencoder* (a weak dictionary learning algorithm) to identify better candidates for EREs in small transformer models. These features represent distinct patterns within the model and are expressed as linear combinations of neuron activations.

In their study of a transformer language model, Bricken et al. decompose a layer with 512 neurons into more than 4,000 features by training sparse autoencoders on multilayer perceptron activations derived from 8 billion data points. Each feature captures a unique concept, such as DNA sequences, legal language, HTTP requests, Hebrew text, or nutrition statements. To evaluate the interpretability of these features—that is, the degree to which humans can understand them—the authors conduct an assessment with a blinded human evaluator (Fig. 3.4). The results show that features are substantially more interpretable than individual neurons, revealing functional properties that remain opaque at the neuron level.

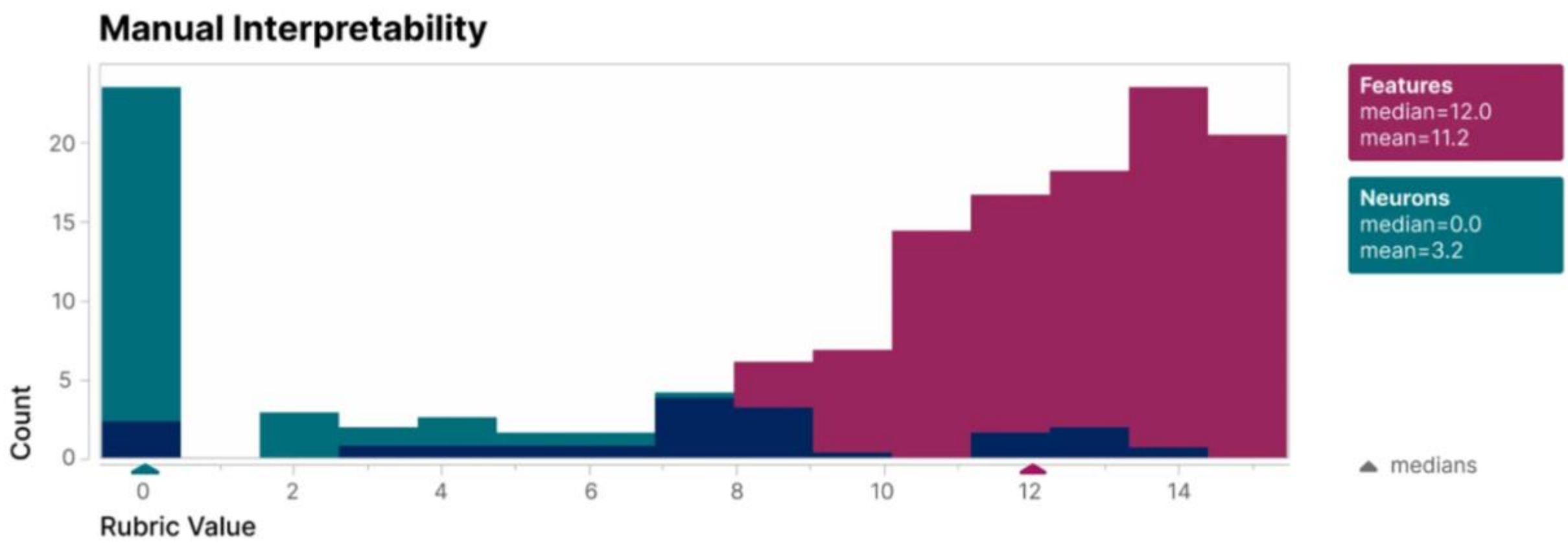

**Fig. 3.4** Interpretability scores of the identified features (red) compared to neurons (teal) (reproduced from Bricken et al., 2023b).

The authors also perform "autointerpretability" tests, in which a language model generates concise descriptions of the small model's features. Another model then evaluates these descriptions by attempting to predict each feature's activations from its textual description. Here too, the features outperform neurons, demonstrating coherent, stable interpretability and a measurable impact on model behavior.

Decomposing the language model into features provides a targeted way of guiding model behavior, as activating specific features produces predictable downstream effects. The researchers also introduce a "knob" to adjust the resolution at which the model is viewed by varying the number of learned features. A smaller set yields a coarse but clear representation, whereas a larger set reveals more refined and subtle properties. Notably, the learned features have proven to generalize across a range of models, highlighting the broader applicability of this explanatory approach.

Overall, evidence from these case studies supports the idea that a coordinated, systematic research program aimed at uncovering the mechanistic organization of DNNs can illuminate how such systems function at multiple structural levels. Mechanistic AI explanations grounded in decomposition and localization can reveal previously unknown EREs in opaque models, which might remain obscured with uncoordinated individual explainability methods.

# 4 Are Deep Neural Networks Genuine Mechanisms?

Several objections from orthodox mechanists challenge the characterization of DNNs as mechanisms, questioning whether these models satisfy the criteria for genuine mechanisms. Despite possible skepticism, I argue that the neomechanistic discovery strategies of decomposition, localization, and recomposition can nevertheless help researchers at least partially open the deep learning black box.

A key standard in mechanistic explanation is the requirement of *completeness*: causal models must explicitly detail all causally relevant parts and operations without gaps or placeholders (Craver, 2007). A fully adequate mechanistic explanation must specify structural details at every relevant level, identifying the components and activities that contribute to concrete computations. This demand appears to conflict with the abstractness and medium-independence characteristic of computational explanations (Coelho Mollo, 2018; Haimovici, 2013). Under this stricter lens, AI models such as DNNs may seem not to qualify as genuine mechanisms, since they typically consist of abstract, formal specifications of computation that lack implementation details. In most cases, DNNs are simulated through matrix operations rather than realized on physical nodes (except in rare cases involving one-to-one mapping, such as on neuromorphic hardware, e.g., Schuman et al., 2022). Satisfying full mechanistic adequacy would thus require additional *instantiation blueprints* (Miłkowski, 2014).

Even so, there are compelling reasons to treat DNNs as mechanisms for explanatory purposes. Although a full investigation of the ontic status of AI models lies beyond the scope of this chapter, I highlight two arguments supporting the idea that DNNs can be meaningfully interpreted through a mechanistic approach.

First, DNNs maintain their mechanistic status through weak *structural constraints* typical of functional explanations, in which structural properties must realize a specified functional characterization (Piccinini, 2015). Piccinini and Craver (2011, p. 302) state, “The functional properties of black boxes are specified in terms of their inputs and outputs (plus the algorithm, perhaps), independently of their physical implementation.” DNNs exemplify this principle, as their inputs, outputs, and mapping algorithms are defined without reference to any particular physical substrate.

Nonetheless, the functional properties of DNNs impose limits on the structural components, since implementing an algorithm requires the system to preserve certain degrees of freedom. Consequently, functional analysis in AI explains how the algorithm, training data, and network architecture determine the structural requirements for computation. When the system's functional components are properly organized and can reliably differentiate among computational vehicles, the same computation may be realized in diverse physical media—mechanical, electromechanical, electronic, or magnetic—without depending on specific physical details.

On this view, deep learning models can be regarded as mathematically defined systems that approximate the behavior of concrete, physically instantiated systems. Their functional properties specify the degrees of freedom that a physical system must exhibit to perform computations. Because these models are not causally complete in the sense required for full-blown mechanistic explanation, their functional analyses should be understood as "mechanism sketches," in which certain structural details are intentionally omitted (Piccinini & Craver, 2011). This level of characterization remains sufficient for XAI research, where fine-grained physical information is often unnecessary for evaluating or improving algorithmic processes.[3] As case studies from OpenAI and Anthropic demonstrate, mechanistic decomposition of AI systems can proceed productively at the functional level without requiring a complete account of every aspect of computation in the underlying physical system.

Second, deep learning models can be understood as *teleofunctional mechanisms*, or simply functional mechanisms, defined by their "purposes" or "ends." Computing systems are designed to compute: they manipulate variables according to rules sensitive to the properties of those variables (Coelho Mollo, 2018; Piccinini, 2015). Deep learning systems satisfy this characterization by performing digital computations through the manipulation of digits and strings of digits—ultimately realized as voltage intervals but represented within the model as numerical vectors transformed through matrix operations.

DNNs operate with medium-independent vehicles (the entities of a mechanism) and the manipulations these vehicles undergo (activities) according to mappings from inputs to outputs determined by a transition function learned during training. These systems consist of organized components that perform specific functions, and their coordinated activities constitute the teleofunctional capacities of deep learning models. Provided that the physical implementation affords the necessary degrees of freedom and avoids miscomputation, the system's abilities are defined by the functional organization of these components rather than their physical makeup.

[3] Exceptions arise when the physical medium affects computational performance—for example, running a large language model on a smartphone, or using inadequate processing hardware for image recognition in autonomous vehicle.

Assessing whether a DNN fulfills its teleological function therefore requires decomposing the model into its components to understand their contributions.

This reasoning extends naturally to mathematically defined computing systems, which stand to concrete implementations roughly as geometric triangles stand to physical triangular objects:

> A similar notion of functional mechanism applies to computing systems that are defined purely mathematically, such as (unimplemented) Turing machines. Turing machines consist of a tape divided into squares and a processing device. The tape and processing device are explicitly defined as spatiotemporal components. They have functions (storing letters; moving along the tape; reading, erasing, and writing letters on the tape) and an organization (the processing device moves along the tape one square at a time, etc.). Finally, the organized activities of their components explain the computations they perform (Piccinini, 2015, pp. 119–120).

Consequently, decomposing teleofunctional mechanisms, even when the target system is defined purely mathematically, yields an elliptical mechanistic explanation of the system's capacities.

To summarize, the core idea of applying the mechanistic approach to deep learning is that, despite skepticism about the ontic status of DNNs, neomechanistic discovery strategies of decomposition, localization, and recomposition remain effective tools for investigating their internal organization. When combined with application-specific analytical explainability techniques, mechanistic analysis enables researchers to identify functionally relevant components and clarify their roles within the system, thereby yielding EREs that support meaningful explanation.

# 5 Epistemic Relevance of the Mechanistic Approach

While the relevance of the mechanistic approach for XAI has been explored in both technical and philosophical literature, this paper advances the discussion by examining the specific epistemic advantages and limitations of mechanistic explanation and situating them within selected philosophical debates about XAI.

## 5.1 Epistemic Advantages

First, mechanistic decomposition of AI systems coheres with the interpretability criteria articulated by Doran et al. (2017). In their framework, an *interpretable system* allows users not only to see but also to study and understand how inputs are mathematically mapped to outputs, thereby helping to "probe the mechanisms of ML systems." Regression models, support vector machines, decision trees, ANOVAs, and clustering models serve as typical examples of such interpretable systems. Although deep neural networks present significant interpretability challenges due to their autonomous feature learning and non-linear transformations, mechanistic decomposition offers a promising avenue for examining their internal dynamics. By isolating human-understandable functional components, this approach allows technically proficient stakeholders to analyze the algorithmic mechanisms that generate a model's behavior.

Second, mechanistic decomposition provides valuable insight into the internal mappings of AI systems, thereby enhancing structural transparency. Creel (2020) defines *structural transparency* as understanding how an algorithm is realized in code, which involves knowledge of its subcomponents and their relationships, typically gained by analyzing system interactions. This understanding enables the modeling of system structure and behavior through tools such as code maps, flowcharts, and diagrams, which closely align with mechanistic principles. Creel is, however, skeptical that such decomposition can meaningfully reduce the structural opacity of DNNs, noting that "when the functional units of the program are tiny, simple, and numerous, as are the neurons of a deep neural network, a subcomponent map would prove insufficient" (2020, pp. 19–20).

Nevertheless, recent advancements in XAI suggest that deeper insight into model functionality can be achieved by identifying units of analysis beyond individual neurons. Examples include neuron circuits (as in Cammarata et al., 2021), and patterns of neuron activations (as in Bricken et al., 2023a, 2023b), uncovered in mechanism discovery processes through techniques such as feature visualization and dictionary learning. While decomposing complex neural networks into functionally meaningful components may leave some neuron-level operations obscured, the mechanistic approach can nonetheless substantially mitigate structural opacity.

Third, the mechanistic explanatory strategy can yield highly generalizable and *counterfactual* explanations of AI decision making. Drawing on Woodward's (2003) interventionist account of causality, Buijsman (2022) argues that effective AI explanations should demonstrate how outcomes would differ under counterfactual conditions. Counterfactual explanations compare actual model outputs with hypothetical alternatives to identify relevant input–output dependencies. Buijsman contends that such contrasts are explanatory when they reveal generalizable correlations across a range of "what-if-things-had-been-different" scenarios. Mechanistic explanations naturally satisfy this requirement by identifying rules and dependencies that generalize across inputs and architectures. In doing so, they meet Buijsman's criterion of providing "reasonable generalizations" that address the counterfactual questions relevant to understanding model behavior.

Although mechanistic approaches typically focus on actual causal processes and may appear to overlook potential counterfactual scenarios, Buijsman's perspective is based on Woodward's (2003) definition of causation, which frames causation itself in counterfactual terms. On this view, causation involves *interventions* that alter one variable without changing others that could affect the outcome, such that "*x* causes *y* if an intervention on *x* changes the value of *y*" (Buijsman, 2022, p. 566). This conception integrates causal–mechanistic reasoning directly into counterfactual evaluation, suggesting that mechanistic considerations can indeed fulfill Buijsman's criteria. In practice, researchers can leverage mechanistic knowledge to explore such scenarios—for example, by perturbing input images to assess network resilience or by evaluating the effects of ablating specific parts of a model.

Finally, Tomsett et al. (2018) argue that the explainability of a machine learning system should be assessed relative to specific stakeholders and tasks. The *black box problem* arises partly because developers struggle to explain system behavior in terms of its learned parameters. However, other stakeholders may require different forms of explanation to meet their respective needs (Zednik, 2021). *Stakeholders* range from developers responsible for engineering and maintaining the system to end users concerned with fairness and usability, each requiring tailored forms of understanding (Hind, 2019; Kasirzadeh, 2021; Tomsett et al., 2018).[4] Recognizing these varied epistemic needs, particularly concerning complexity and domain-specific language, is essential for supporting stakeholders in fulfilling their respective roles within an AI ecosystem.

The adoption of the mechanistic strategy for XAI, which emphasizes multilevel hierarchies of mechanisms, can address these diverse needs by offering understand- able explanations at multiple levels of abstraction. Mechanistic reasoning may allow AI developers to explore nested hierarchical structures to identify key causal variables, such as learnable parameters and representational components, that transform inputs into outputs. Consequently, detailed lower-level explanations can help them identify errors or enhance performance, addressing their epistemic needs for model control, manipulation, and prediction. By contrast, empirical findings from Ribeiro et al. (2016) indicate that while machine learning experts can navigate such complex explanations, laypersons prefer simplified accounts that reduce models to a small set of weighted features. Accordingly, higher-level simplified descriptions of mechanisms, functional overviews, or abstract schemata may better serve end users, who primarily seek intelligible information that supports trust and practical decision making.

[4] Specific understanding of stakeholders is brought forward by the EU AI Act, which focuses on deployers: natural or legal persons, including a public authority, agency or other body, using an AI system under its authority, except where the AI system is used in the course of a personal non-professional activity (European Union, 2024).

This also aligns with Buijsman's (2022) call for abstract variables in explanations to increase generality and reduce cognitive load. A mechanistic multilevel approach potentially enables researchers to tailor explanations to diverse audiences and may also facilitate compliance with regulatory requirements, such as those imposed by the European Union in the AI Act (European Union, 2024), by adjusting the level of explanatory detail. Recall Anthropic's research on language model decomposition, which involved creating a "knob" to adjust the model's visibility resolution and experimenting with the number of features learned (Bricken et al., 2023a). In this way, while the mechanistic explanatory strategy seeks to illuminate the internal workings of AI models, it also provides a framework for crafting human-grounded explanations aligned with the epistemic requirements of diverse stakeholders.

Overall, to properly assess the impact of such multilevel approaches, further empirical studies involving multiple AI stakeholders are essential for understanding how explanations are perceived and interpreted by diverse audiences (cf. Dorsch & Moll, 2026).

## 5.2 Epistemic Limitations

While a theoretically grounded mechanistic exploratory strategy holds promise for XAI in terms of its considerable epistemic advantages, its overall utility is constrained by certain epistemic limitations.

First, adopting Doran et al.'s (2017) XAI typology, mechanistic explanations can enhance interpretability by revealing the inner workings of AI systems, but they do not necessarily improve comprehensibility. *Comprehensibility* involves users making sense of outputs through interpretable symbols like words or visualizations, regardless of the system's internal opacity. "Autointerpretability" techniques, such as those developed by Bricken et al. (2023b) and Schwettmann et al. (2023), generate natural language descriptions of model components to support comprehension. Yet understanding these descriptions depends on users' implicit background knowledge. Although visualization methods can display recognizable features in image recognition systems, XAI techniques frequently identify subtle or highly abstract features that may elude human understanding (Buckner, 2019; Zednik, 2019). Thus, while mechanistic AI explanations should ideally present mechanisms in human-understandable terms, the statistical nature of deep learning frequently diverges from intuitive concepts.

Moreover, comprehending such explanations arguably requires a degree of technical proficiency with AI, and this requirement varies across explanatory types. Higher-level explanations that abstract away from fine-grained details likely demand less expertise than lower-level accounts that closely track internal workings. This variability was evident in Ribeiro et al.'s (2016) evaluation of LIME (though not itself a mechanistic case), which relied on trained computer science graduates as participants. Such dependence on technical expertise presents a significant challenge, as stakeholders beyond system developers will continue to demand explanations for system behavior even when they lack the necessary background.

Second, Creel's (2020) distinction between different forms of transparency high- lights further limitations of the mechanistic approach. Although mechanistic analysis can support structural transparency, it may fall short of achieving algorithmic and run transparency. *Algorithmic transparency* concerns understanding the high-level logical rules that govern system-wide transformations, which is not fully secured by decomposing a model into its constituent functions. *Run transparency*, on the other hand, requires knowledge of how specific program operations unfold in practice, including hardware-level behavior and the processing of particular inputs. It involves observing how software executes on concrete hardware with real data. Because the mechanistic explanatory strategy focuses on abstract, mathematically specified models and their degrees of freedom, it typically does not capture artifacts arising from real-time interactions between software and hardware, unexpected input distributions, or the consequences of compiling code into machine instructions.

Finally, there is the issue of *complexity*. Whereas classical computer systems are relatively transparent, deep learning models are considered black boxes due to the complex interdependencies among millions of parameters composing their internal states. This complexity enables neural networks to excel at problem solving but makes it difficult to articulate their causal–mechanical structure. As Kostić (2023) observes, the opacity arising from such functional complexity renders the mechanistic ideal—complete knowledge of components, activities, and organization—practically unattainable from the outset. Even if not epistemically impossible, meeting this standard is practically challenging given current engineering methods, resource constraints, and the accelerating demand for explanations in real-world AI systems. In practice, mechanistic analysis may be limited to small, localized subsystems, much like the granularity observed in contemporary neuroscience.

Complex DNNs may resemble *non-decomposable systems*, in which each component's behavior is heavily influenced by its interactions with many others (cf. Rathkopf, 2018). Although decomposition helps manage the complexity of representing every element, thereby mitigating combinatorial explosion, it may yield explanations that are narrow in scope and applicable only to specific subsystems or simplified toy models. As Creel (2020) observes, while some input–output paths may be tractable, fully understanding all subcomponents can be prohibitively complex.

This has motivated efforts to scale microscopic insights from mechanistic interpretability research to larger models.[5] However, skepticism about such scalability persists due to computational constraints, financial cost, and unresolved methodological challenges (Casper, 2023; Greenblatt et al., 2023; Nanda, 2023). Evidence from small-scale models does not guarantee that real-world DNNs can be decomposed in a way that supports comprehensive mechanistic understanding. When part–whole decomposition is infeasible, approaches that are fueled, not hampered, by a system's complexity—such as network science and topological explanations (e.g., Kostić, 2022; Rathkopf, 2018)—offer promising alternatives.

## 6 Conclusions

The mechanistic explanatory strategy for XAI focuses on identifying the mechanisms that drive automated decision making. For deep neural networks, this involves isolating functionally relevant components—such as neurons, layers, circuits, or activation patterns—and clarifying their roles through the heuristic discovery strategies of decomposition, localization, and recomposition. Early research suggests that such a coordinated, systematic approach to analyzing a model's functional organization can reveal epistemically relevant elements that simpler explainability techniques may overlook, thereby enabling more effective forms of explanation. Supported by real- world examples from image recognition and language modeling, this philosophical analysis thus highlights the value of mechanistic reasoning for advancing XAI.

The mechanistic approach also offers significant epistemic benefits. It enhances interpretability and structural transparency, supports the construction of counterfactual and highly generalizable explanations, and aids in prediction and system manipulation by clarifying internal causal dynamics. Its multilevel, hierarchical structure allows explanations to be tailored to diverse stakeholder needs, making complex AI systems more accessible and manageable.

[5] For example, in OpenAI's study of curve detectors, researchers reverse-engineered how the first four layers of InceptionV1 progressively build toward curve detectors in the fifth layer, analyzing a family of 10 such neurons (Cammarata et al., 2021). Yet the full InceptionV1 model comprises 22 layers (27 including pooling) and between 5 and 6 million parameters. Likewise, in an Anthropic study, researchers examined a deliberately small, one-layer transformer with a 512-neuron MLP (Bricken et al., 2023a). By contrast, GPT-3 (the predecessor of GPT versions used in ChatGPT) contains 175 billion parameters across 96 layers.

Deepening our understanding of AI mechanisms and their causal organization can improve performance assessment and reveal avenues for refinement. As such, further development of mechanistic frameworks for XAI may be essential for addressing persistent challenges of trust and transparency, especially in high-stakes algorithmic decision making.

However, despite its promise, the mechanistic strategy faces certain epistemic limitations. These include ambiguous contributions to algorithmic and run transparency, limited comprehensibility due to the absence of suitable concepts, the need for recipients to possess technical proficiency in AI, and the possible practical challenges of decomposing complex, real-world systems that are not apparent in simplified toy models.

Several areas of future research follow from these considerations. First, it is important to evaluate the applicability and scalability of the mechanistic approach beyond toy models and toward more complex AI systems. Second, empirical investigation into how different stakeholders comprehend mechanistic explanations, complemented by a suitable epistemology of understanding, would be a valuable next step. Finally, given the limitations identified, it is essential to explore additional philosophically informed explanatory strategies, particularly those that can capitalize on, rather than be hindered by, system complexity and that can address other forms of opacity.

**Acknowledgements** I would like to thank Marcin Miłkowski, Dimitri Coelho Mollo, Lena Kästner, Barnaby Crook, Kristian González Barman, and John Dorsch for their constructive comments that helped improve this manuscript.

**Funding** This work was supported by the National Science Centre, Poland, under PRELUDIUM grant no. 2023/49/N/HS1/02461, and by the Polish National Agency for Academic Exchange under NAWA STER project no. BPI/STE/2021/1/00030/U/00001.

## References

Bechtel, W., & Abrahamsen, A. (2005). Explanation: a mechanist alternative. *Studies in History and Philosophy of Science Part C: Studies in History and Philosophy of Biological and Biomedical Sciences*, *36*(2), 421–441. https://doi.org/10.1016/j.shpsc.2005.03.010

Bechtel, W., & Abrahamsen, A. A. (2013). Thinking dynamically about biological mechanisms: Networks of coupled oscillators. *Foundations of Science*, *18*, 707–723. https://doi.org/10.1007/s10699-012-9301-z

Bechtel, W., & Richardson, R. C. (2010). *Discovering complexity: Decomposition and localization as strategies in scientific research* (2nd ed.). MIT Press/Bradford Books. https://doi.org/10.7551/mitpress/8328.001.0001

Beisbart, C., & Räz, T. (2022). Philosophy of science at sea: Clarifying the interpretability of machine learning. *Philosophy Compass, 17*(12), e12830. https://doi.org/10.1111/phc3.12830

Bricken, T., Templeton, A., Batson, J., Olah, C., Henighan, T., Carter, S., Hume, T., Burke, J. E., McLean, B., Nguyen, K., Tamkin, A., Joseph, N., Maxwell, T., Schiefer, N., Kravec, S., Wu, Y., Lasenby, R., Askell, A., Denison, C., … Chen, B. (2023, October 4). *Towards monosemanticity: Decomposing language models with dictionary learning*. Anthropic. https://transformer-circuits.pub/2023/monosemantic-features/index.html

Bricken, T., Templeton, A., Batson, J., Olah, C., Henighan, T., Carter, S., Hume, T., Burke, J. E., McLean, B., Nguyen, K., Tamkin, A., Joseph, N., Maxwell, T., Schiefer, N., Kravec, S., Wu, Y., Lasenby, R., Askell, A., Denison, C., … Chen, B. (2023, October 5). *Decomposing language models into understandable components*. Anthropic. https://www.anthropic.com/index/decomposing-language-models-into-understandable-components

Buckner, C. (2019). Deep learning: A philosophical introduction. *Philosophy Compass, 14*(10), e12625. https://doi.org/10.1111/phc3.12625

Buijsman, S. (2022). Defining explanation and explanatory depth in XAI. *Minds and Machines*, *32*(3), 563–584. https://doi.org/10.1007/s11023-022-09607-9

Cammarata, N., Goh, G., Carter, S., Voss, C., Schubert, L., & Olah, C. (2021). Curve circuits. *Distill*. https://doi.org/10.23915/distill.00024.006

Casper, S. (2023, February 17). EIS VI: Critiques of mechanistic interpretability work in AI safety. *AI Alignment Forum*. https://www.alignmentforum.org/posts/wt7HXaCWzuKQipqz3/eis-vi-critiques-of-mechanistic-interpretability-work-in-ai

Chan, L., Garriga-Alonso, A., Goldowsky-Dill, N., Greenblatt, R., Nitishinskaya, J., Radhakrishnan, A., Shlegeris, B., & Thomas, N. (2022, December 3). Causal scrubbing: A method for rigorously testing interpretability hypotheses. *AI Alignment Forum*. https://www.alignmentforum.org/posts/JvZhhzycHu2Yd57RN/causal-scrubbing-a-method-for-rigorously-testing

Christiano, P. (2022, November 25). Mechanistic anomaly detection and ELK. *AI Alignment*. https://ai-alignment.com/mechanistic-anomaly-detection-and-elk-fb84f4c6d0dc

Coelho Mollo, D. (2018). Functional individuation, mechanistic implementation: The proper way of seeing the mechanistic view of concrete computation. *Synthese, 195*, 3477–3497. https://doi.org/10.1007/s11229-017-1380-5

Conmy, A., Mavor-Parker, A.N., Lynch, A., Heimersheim, S., & Garriga-Alonso, A. (2023). Towards automated circuit discovery for mechanistic interpretability. *arXiv*. https://doi.org/10.48550/arXiv.2304.14997

Craver, C. F. (2007). *Explaining the brain: Mechanisms and the mosaic unity of neuroscience.* Clarendon Press.

Creel, K. A. (2020). Transparency in complex computational systems. *Philosophy of Science, 87*(4), 568–589. https://doi.org/10.1086/709729

Cunningham, H., Ewart, A., Riggs, L., Huben, R., & Sharkey, L. (2023). Sparse autoencoders find highly interpretable features in language models. *arXiv*. https://doi.org/10.48550/arXiv.2309.08600

Doran, D., Schulz, S., & Besold, T. R. (2017). What does explainable AI really mean? A new conceptualization of perspectives. In T. R. Besold & O. Kutz (Eds.), *Proceedings of the First International Workshop on Comprehensibility and Explanation in AI and ML 2017* (CEUR Workshop Proceedings, Vol. 2071, pp. 1–8). https://ceur-ws.org/Vol-2071/CExAIIA_2017_paper_2.pdf

Dorsch, J., & Moll, M. (2026). Explainable and human-grounded AI for decision support systems: The theory of epistemic quasi-partnerships. In V. C. Müller, L. Dung, G. Löhr, & A. Rumana (Eds.), *Philosophy of artificial intelligence: The state of the art* (pp. 83–103, Synthese Library, Vol. 533). Springer. https://doi.org/10.1007/978-3-032-10073-3_5

Elhage, N., Nanda, N., Olsson, C., Henighan, T., Joseph, N., Mann, B., Askell, A., Bai, Y., Chen, A., Conerly, T., DasSarma, N., Drain, D., Ganguli, D., Hatfield-Dodds, Z., Hernandez, D., Jones, A., Kernion, J., Lovitt, L., … Olah, C. (2021, December 22). A mathematical framework for transformer circuits. *Transformer Circuits Thread*. https://transformer-circuits.pub/2021/framework/index.html

Erasmus, A., Brunet, T.D.P., & Fisher, E. (2021). What is interpretability? *Philosophy & Technology*, *34,* 833–862. https://doi.org/10.1007/s13347-020-00435-2

European Union. (2024). Regulation (EU) 2024/1689 of the European Parliament and of the Council of 13 June 2024 laying down harmonised rules on artificial intelligence and amending certain Union legislative acts (Artificial Intelligence Act). *Official Journal of the European Union*, L 1689. https://eur-lex.europa.eu/eli/reg/2024/1689/oj

Glennan, S. S. (1996). Mechanisms and the nature of causation. *Erkenntnis, 44,* 49–71. https://doi.org/10.1007/BF00172853

Glennan, S. S. (2017). *The new mechanical philosophy.* Oxford University Press.

Greenblatt, R., Nanda, N., Buck, & Shlegeris, B., & Habryka, O. (2023, December 1). How useful is mechanistic interpretability? *LessWrong.* https://www.lesswrong.com/posts/tEPHGZAb63dfq2v8n/how-useful-is-mechanistic-interpretability

Haimovici, S. (2013). A problem for the mechanistic account of computation. *Journal of Cognitive Science*, *14*(2), 151–181. http://doi.org/10.17791/jcs.2013.14.2.151

Hind, M. (2019). Explaining explainable AI. *XRDS: Crossroads, The ACM Magazine for Students*, *25,* 16–19. https://doi.org/10.1145/3313096

Humphreys, P. (2009). The philosophical novelty of computer simulation methods. *Synthese, 169*(3), 615–626. https://doi.org/10.1007/s11229-008-9435-2

Illari, P., & Williamson, J. (2012). What is a mechanism? Thinking about mechanisms across the sciences. *European Journal for Philosophy of Science, 2*(1), 119–135. http://dx.doi.org/10.1007/s13194-011-0038-2
Karmakharm, T. (2018). Image classification with DIGITS [Presentation]. NVIDIA Deep Learning Institute. https://rse.shef.ac.uk/assets/slides/2018-07-19-dl-cv/image-classification.pdf
Kasirzadeh, A. (2021). Reasons, values, stakeholders: A philosophical framework for explainable artificial intelligence. In *Proceedings of the 2021 ACM Conference on Fairness, Accountability, and Transparency*. Association for Computing Machinery. https://doi.org/10.1145/3442188.3445866
Kästner, L., & Crook, B. (2024). Explaining AI through mechanistic interpretability. *European Journal for Philosophy of Science*, *14,* 52. https://doi.org/10.1007/s13194-024-00614-4
Kostić, D. (2022). Topological explanations: An opinionated appraisal. In I. Lawler, E. Shech, & K. Khalifa (Eds.), *Scientific understanding and representation: Mathematical modeling in the life and physical sciences* (pp. 96–115). Routledge. https://doi.org/10.4324/9781003202905-9
Kostić, D. (2023). *Pragmatics of explainability relevance in XAI* [Manuscript].
Kostić, D., & Halffman, W. (2023). Mapping explanatory language in neuroscience. *Synthese*, *202*, 112. https://doi.org/10.1007/s11229-023-04329-6
Linardatos, P., Papastefanopoulos, V., & Kotsiantis, S.B. (2021). Explainable AI: A review of machine learning interpretability methods. *Entropy, 23*(1), 18. https://doi.org/10.3390/e23010018
Lipton, Z. C. (2018). The mythos of model interpretability. *Queue, 16*(3), 31–57. https://doi.org/10.1145/3236386.3241340
Machamer, P. K., Darden, L., & Craver, C. F. (2000). Thinking about mechanisms. *Philosophy of Science, 67*(1), 1–25. https://doi.org/10.1086/392759
Marrow, S., Michaud, E. J., & Hoel, E. (2020). Examining the causal structures of deep neural networks using information theory. *Entropy*, *22*(12), 1429. https://doi.org/10.3390/e22121429
Miłkowski, M. (2014). Computational mechanisms and models of computation. *Philosophia Scientiæ, 18*(3), 215–228. https://doi.org/10.4000/philosophiascientiae.1019
Miller, T. (2019). Explanation in artificial intelligence: Insights from the social sciences. *Artificial Intelligence, 267*, 1–38. https://doi.org/10.1016/j.artint.2018.07.007
Miller, T., Howe, P., & Sonenberg, L. (2017). Explainable AI: Beware of inmates running the asylum or: How I learnt to stop worrying and love the social and behavioural sciences. *arXiv*. https://doi.org/10.48550/arXiv.1712.00547
Millière, R., & Buckner, C. (2025). Interventionist methods for interpreting deep neural networks. In G. Piccinini (Ed.), *Neurocognitive foundations of mind* (1st ed., pp. 190–221). Routledge. https://doi.org/10.4324/9781003458531
Mittelstadt, B., Russell, C., & Wachter, S. (2019). Explaining explanations in AI. In *Proceedings of the Conference on Fairness, Accountability, and Transparency* (pp. 279–288). Association for Computing Machinery. https://doi.org/10.1145/3287560.3287574
Molnar, C. (2022). *Interpretable machine learning: A guide for making black box models explainable* (2nd edition). https://christophm.github.io/interpretable-ml-book/
Montavon, G., Samek, W., & Müller, K.-R. (2018). Methods for interpreting and understanding deep neural networks. *Digital Signal Processing, 73*, 1–15. https://doi.org/10.1016/j.dsp.2017.10.011
Nanda, N. (2023, July 6). Concrete open problems in mechanistic interpretability: A technical overview. *Effective Altruism Forum*. https://forum.effectivealtruism.org/posts/EMfLZXvwiEioPWPga/concrete-open-problems-in-mechanistic-interpretability-a
Olah, C. (2022). Mechanistic interpretability, variables, and the importance of interpretable bases. *Transformer Circuit Thread*. https://transformer-circuits.pub/2022/mech-interp-essay/index.html
Olah, C., Mordvintsev, A., & Schubert, L. (2017). Feature visualization. *Distill*. https://doi.org/10.23915/distill.00007
Olsson, C., Elhage, N., Nanda, N., Joseph, N., DasSarma, N., Henighan, T. J., Mann, B., Askell, A., Bai, Y., Chen, A., Conerly, T., Drain, D., Ganguli, D., Hatfield-Dodds, Z., Hernandez, D., Johnston, S., Jones, A., Kernion, J., Lovitt, L., … Olah, C. (2022). In-context learning and induction heads. *arXiv*. https://doi.org/10.48550/arXiv.2209.11895
Páez, A. (2019). The pragmatic turn in explainable artificial intelligence (XAI). *Minds and Machines, 29,* 441–459. https://doi.org/10.1007/s11023-019-09502-w
Piccinini, G. (2015). *Physical computation: A mechanistic account.* Oxford University Press. https://doi.org/10.1093/acprof:oso/9780199658855.001.0001
Piccinini, G., & Craver, C. (2011). Integrating psychology and neuroscience: Functional analyses as mechanism sketches. *Synthese, 183*, 283–311. https://doi.org/10.1007/s11229-011-9898-4
Qin, Z., Yu, F., Liu, C., & Chen, X. (2018). How convolutional neural networks see the world — A survey of convolutional neural network visualization methods. *Mathematical Foundations of Computing*, *1*(2), 149–180. https://doi.org/10.3934/mfc.2018008

Rathkopf, C. (2018). Network representation and complex systems. *Synthese, 195*, 55–78. https://doi.org/10.1007/s11229-015-0726-0

Ribeiro, M. T., Singh, S., & Guestrin, C. (2016). "Why should I trust you?": Explaining the predictions of any classifier. In *Proceedings of the 22nd ACM SIGKDD International Conference on Knowledge Discovery and Data Mining* (pp. 1135–1144). Association for Computing Machinery. https://doi.org/10.1145/2939672.2939778

Schuman, C. D., Kulkarni, S. R., Parsa, M., Mitchell, J. P., Date, P., & Kay, B. (2022). Opportunities for neuromorphic computing algorithms and applications. *Nature Computational Science, 2*(1), 10–19. https://doi.org/10.1038/s43588-021-00184-y

Schwettmann, S., Shaham, T. R., Materzynska, J., Chowdhury, N., Li, S., Andreas, J., Bau, D., & Torralba, A. (2023). FIND: A function description benchmark for evaluating interpretability methods. *arXiv*. https://doi.org/10.48550/arXiv.2309.03886

Schyns, P. G., Snoek, L., & Daube, C. (2022). Degrees of algorithmic equivalence between the brain and its DNN models. *Trends in Cognitive Sciences*, *26*, 1090–1102. https://doi.org/10.1016/j.tics.2022.09.003

Shahriar, N. (2023, February 1). What is convolutional neural network — CNN (deep learning). *Medium.* https://nafizshahriar.medium.com/what-is-convolutional-neural-network-cnn-deep-learning-b3921bdd82d5

Tomsett, R., Braines, D., Harborne, D., Preece, A., & Chakraborty, S. (2018). Interpretable to whom? A role-based model for analyzing interpretable machine learning systems. *arXiv*. https://doi.org/10.48550/arXiv.1806.07552

Watson, D. S., & Floridi, L. (2020). The explanation game: A formal framework for interpretable machine learning. *Synthese, 198*(10), 9211–9242. https://doi.org/10.1007/s11229-020-02629-9

Woodward, J. (2003). *Making things happen: A theory of causal explanation*. Oxford University Press.

Wright, C., & Bechtel, W. (2007). Mechanisms and psychological explanation. In P. Thagard (Ed.), *Philosophy of psychology and cognitive science* (pp. 31–79). Elsevier.

Yosinski, J., Clune, J., Nguyen, A. M., Fuchs, T. J., & Lipson, H. (2015). Understanding neural networks through deep visualization. *arXiv*. https://doi.org/10.48550/arXiv.1506.06579

Zednik, C. (2021). Solving the black box problem: A normative framework for explainable artificial intelligence. *Philosophy & Technology, 34*(2), 265–288. https://doi.org/10.1007/s13347-019-00382-7

Zerilli, J., Knott, A., MacLaurin, J., & Gavaghan, C. (2019). Transparency in algorithmic and human decision-making: Is there a double standard? *Philosophy & Technology*, *32*, 661–683. https://doi.org/10.1007/s13347-018-0330-6